\documentclass[conference,11pt]{IEEEtran}
\IEEEoverridecommandlockouts
\usepackage{amsmath,amssymb,amsfonts}
\usepackage{algorithm}
\usepackage{algpseudocode}

\usepackage{graphicx}
\usepackage[draft]{hyperref}
\usepackage{textcomp}
\usepackage{cite}
\usepackage{booktabs}
\usepackage{multirow}
\usepackage{paralist}

\usepackage{array}
\def\BibTeX{{\rm B\kern-.05em{\sc i\kern-.025em b}\kern-.08em
   T\kern-.1667em\lower.7ex\hbox{E}\kern-.125emX}}
\newcommand{\nix}[1]{}  

\begin{document}
\title{{Early Detection of Acute Myeloid Leukemia (AML) Using YOLOv12 Deep Learning Model}}
\author{\IEEEauthorblockN{Enas E. Ahmed}
\IEEEauthorblockA{\textit{Biomedical and Systems Dept.} \\
\textit{Faculty of Engineering} \\
\textit{Cairo  University}\\
Giza, Egypt}
\and
\IEEEauthorblockN{Salah A. Aly}
\IEEEauthorblockA{\textit{CS \& Math Branch} \\
\textit{Faculty of Science} \\
\textit{Fayoum University}\\
Fayoum, Egypt }
\and
\IEEEauthorblockN{Mayar Moner}
\IEEEauthorblockA{\textit{Biotechnology Dept.} \\
\textit{Faculty of Agriculture} \\
\textit{Cairo University}\\
Giza, Egypt }
}
\maketitle              
\smallskip
\begin{abstract}
Acute Myeloid Leukemia (AML) is one of the most life-threatening type of blood cancers, and its accurate classification is considered and remains a challenging task due to the visual similarity between various cell types.    This study addresses the classification of the multiclasses of AML cells Utilizing YOLOv12 deep learning model. We applied two segmentation approaches based on cell and nucleus features, using Hue channel and Otsu thresholding techniques to preprocess the images prior to classification. Our experiments demonstrate that YOLOv12 with Otsu thresholding on cell-based segmentation achieved the highest level of validation and test accuracy, both reaching 99.3\%. 
\end{abstract}
\textbf{\emph{Keywords}:} Acute Myeloid Leukemia, AML, disease detection, Deep Learning Models, Transfer Learning.
\section{Introduction}
Leukemia originates in the bone marrow—the central site of hematopoiesis—and can occur at any age, though it is more commonly diagnosed in individuals under 15 years old and those over 55. Leukemia is broadly classified into four main types based on the speed of disease progression (acute vs. chronic) and the lineage of affected cells (myeloid vs. lymphoid): acute myeloid leukemia (AML), acute lymphoblastic leukemia (ALL), chronic myeloid leukemia (CML), and chronic lymphocytic leukemia (CLL). Acute leukemias involve immature, rapidly dividing cells and progress quickly, while chronic leukemias affect more mature cells and develop more slowly\cite{zare2024automatic,SEER_AML}.

AML is the most prevalent form of acute leukemia in adults and is characterized by the rapid proliferation and accumulation of immature myeloid precursors—referred to as myeloblasts—in the bone marrow and peripheral blood~\cite{SEER_AML}. It accounts for about 80\% of all leukemia cases.These abnormal cells typically exhibit a high nuclear-to-cytoplasmic ratio, prominent nucleoli, and atypical morphology. The development of AML is often attributed to genetic mutations, chromosomal abnormalities, and environmental exposures such as smoking, chemotherapy, and radiation\cite{vakiti2024acute}.

This study utilizes a processed dataset consisting of cells that are essential for understanding the progression and diagnosis of acute myeloid leukemia (AML). The dataset focuses on five key blood cell types relevant to AML classification: myeloblasts, segmented neutrophils, basophils, monocytes, and erythroblasts.

Myeloid cells are cells derived from hematopoietic stem cells in the bone marrow and play essential roles in both immunological defense and blood cell development. The myeloid lineage includes both mature white blood cells—like neutrophils, eosinophils, basophils, and monocytes—as well as their early forms (immature), such as myeloblasts~\cite{davis2014leukemia}.

\begin{figure*}[t] 
\centering
\includegraphics[width=1\linewidth, height=6cm]{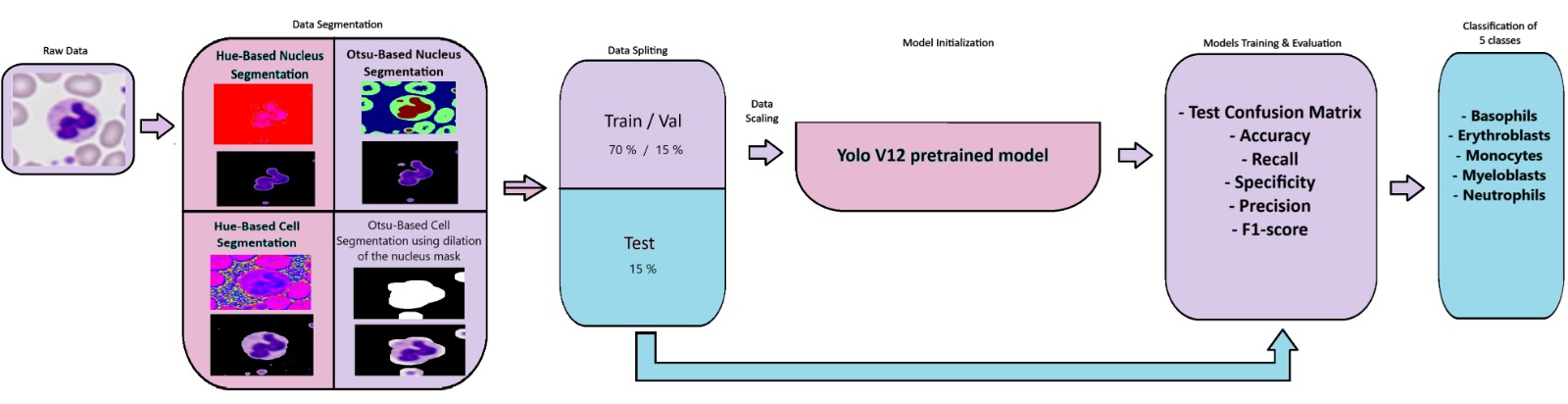}
\caption{The implementation process of our proposed framework, showing the key stages from data preprocessing to final classification.}
\label{implementation}
\end{figure*}

Among the selected cell types, myeloblasts are immature myeloid precursors whose abnormal proliferation is a hallmark of AML. These cells are typically characterized by a large, round, or oval nucleus and fine chromatin~\cite{NAEIM2018293}. In contrast, segmented neutrophils, the mature form of granulocytes, indicate proper myeloid differentiation; their count or morphology deviations can indicate leukemic  transformation~\cite{KHANNAGUPTA2018321}. Basophils, though rare, are granulocytes that originate from the same progenitor as myeloblasts. Their granules often obscure the nucleus, and elevated levels are sometimes observed in AML cases~\cite{papadakis2024approach}. Monocytes are large cells featuring kidney-shaped nuclei involved in immunological surveillance, and increased monocyte counts are often associated with monocytic subtypes of AML. Myeloid cells and erythroblasts share a common progenitor. Their increased presence can be observed in some AML variants, and they typically demonstrate dense chromatin and a circular morphology~\cite{vakiti2024acute}. By incorporating immature and mature myeloid cells, the model aims to capture a biologically meaningful spectrum of hematopoiesis relevant to AML classification.

Despite recent progress, current research in AML classification faces several key limitations. Some have reliability and reproducibility issues in the DL models, and some have issues with the performance and the dataset:
\begin{inparaenum}
\item Limited availability of well-annotated histopathological datasets for training and validation.
\item Model evaluation loses reliability when the test set is used for anything other than assessing final performance. Hyperparameter tuning should be based solely on a separate validation set. Without an independent test set, as seen in some studies, performance results may be overly optimistic or misleading \cite{baig2022detecting,bairaboina2023ghost}.
\item Imbalanced representation of some immature WBC types—such as myeloblasts and promyelocytes—reduces the model's ability to learn their distinct features effectively, leading to lower classification performance~\cite{elhassan2024cae}.
\end{inparaenum}
To address the identified challenges in the current research, this study offers the following key contributions:
\begin{compactenum}[i)]
\item Improved model generalization by training on a balanced dataset comprising diverse myeloid cell types related to AML.
\item Application of tuned transfer learning models using histopathological image data to enhance model performance.
\item Increased accuracy in the early detection of AML.
 \item Utilization of YOLOv12, the latest publicly released version as of 2025/02/19~\cite{tian2502yolov12}, for more robust classification.
 \end{compactenum}
This study proposes a comprehensive deep learning workflow for the classification of AML, driven by the necessity for scalable and automated diagnostic tools.  The process comprises segmentation procedures and the implementation of transfer learning utilizing existing models, including YOLOv12, ResNet50, and InspectionResNet50 v2 as we will present in an extended version.

\section{Datasets Description and Data Collections}\label{sec:dataset}
For this research, a balanced and enhanced version of a dataset was utilized, available on \href{https://www.kaggle.com/datasets/sumithsingh/blood-cell-images-for-cancer-detection/data}{Kaggle}. This version consolidates high-quality cellular images from various resources such as The Cancer Imaging Archive (TCIA) and the Core Laboratory at the Hospital Clinic of Barcelona which is publicly available from two primary sources: \href{https://data.mendeley.com/datasets/snkd93bnjr/1}{Mendeley Data} and \href{https://www.kaggle.com/datasets/unclesamulus/blood-cells-image-dataset/data}{Kaggle}, tailored specifically for leukemia detection. The data contributes both normal and abnormal blood cell images. Each image was captured using Wright-Giemsa staining at 100x oil immersion magnification (total magnification of 1000x), ensuring detailed visualization of single-cell morphological features. Images maintain standardized quality with 1024×1024 pixel resolution, 24-bit RGB color, centered cells, and normalized backgrounds—key features for training deep learning (DL) models. The data is distributed as in Table \ref{data_distribution}. It was split into training, validation, and testing.  

This processed dataset focuses on five key blood cell types critical for leukemia research:

\begin{compactenum}[-]
    \item \textbf{Basophils:} characterized by dark purple cytoplasmic granules.
    \item \textbf{Erythroblasts:} immature red blood cells.
    \item \textbf{Monocytes:}  large agranulocytes with kidney-shaped nuclei.
    \item \textbf{Myeloblasts:}  immature white blood cells, hallmark of AML.
    \item \textbf{Segmented Neutrophils:}  mature granulocytes with segmented nuclei.
\end{compactenum}

This curated subset, licensed under CC BY-NC 4.0, is designed for non-commercial use and serves as a reliable resource for developing AI-driven AML detection systems.
\begin{table}[htbp]
\caption{Dataset Distribution for Leukemia-Related Blood Cell Types (70\% Train, 15\% Val, 15\% Test)}
\centering
\begin{tabular}{|p{2cm}|p{1.2cm}|p{1cm}|p{1cm}|p{1cm}|}
\hline
\textbf{Cell Type} & \textbf{No. of Images} & \textbf{Training (70\%)} & \textbf{Validation (15\%)} & \textbf{Testing (15\%)} \\
\hline
Basophils & 1000 & 700 & 150 & 150 \\
Erythroblasts & 1000 & 700 & 150 & 150 \\
Monocytes & 1000 & 700 & 150 & 150 \\
Myeloblasts & 1000 & 700 & 150 & 150 \\
Segmented Neutrophils & 1000 & 700 & 150 & 150 \\
\hline
\textbf{Total} & \textbf{5000} & \textbf{3500} & \textbf{750} & \textbf{750} \\
\hline
\end{tabular}
\label{data_distribution}
\end{table}
\section{Models and Methodologies}\label{sec:methodology}
The methodology of this study is implemented into three phases as illustrated in Fig.~\ref{implementation}. The first phase involves data segmentation, where two image segmentation techniques—Hue channel filtering and Otsu thresholding—were applied. Each segmentation method used for isolating the cells and nuclei independently, resulting in four distinct segmented image datasets. Then, all images were rescaled to match the input size of each pretrained model.  In the second phase, the pretrained models were independently trained and validated on each of the four segmented datasets. 
%
%
\subsection{YOLOv12}\label{yoloV12}
The \textit{You Only Look Once (YOLO)} model changed the way object detection is done by combining all detection steps into a single, end-to-end trainable neural network. Unlike older methods that first locate and then classify objects separately, YOLO treats detection as a single task. It looks at the whole image once and directly predicts both the bounding boxes and their object classes. The image is divided into an \( S \times S \) grid, where each grid cell predicts multiple bounding boxes, their confidence scores, and class probabilities. These predictions are packed into a tensor of shape \( S \times S \times (B \cdot 5 + C) \), allowing for fast and accurate real-time object detection~\cite{redmon2016you}.

\textit{YOLOv12} is the latest model in the YOLO series, developed by Ultralytics. Previous YOLO versions mainly used convolutional neural networks (CNNs) to ensure speed. YOLOv12 improves on this by adding a new component called \textit{area attention}. This mechanism helps the model focus on important parts of the image while keeping the detection process fast. Unlike traditional attention methods that are often too slow for real-time use, area attention balances speed and contextual understanding efficiently. The version used in this study, \textit{YOLOv12-N}, is both fast and accurate. It achieves a mean Average Precision (mAP) of 40.6\% with only 1.64 milliseconds of delay on an NVIDIA T4 GPU. This makes it more accurate than YOLOv10-N and YOLOv11-N by 2.1\% and 1.2\% respectively, while maintaining real-time performance~\cite{tian2502yolov12}.
\subsection{Area Attention Mechanism in YOLOv12}
YOLOv12 includes a special attention method called \textbf{Area Attention}, which is designed to improve how the model focuses on important regions in an image, without slowing down the detection process~\cite{tian2502yolov12}.

Let’s say we have a feature map from the image, represented as:
\[
\mathbf{F} \in \mathbb{R}^{n \times h \times d}
\]
where \( n \) is the number of tokens (e.g., spatial locations), \( h \) is the number of attention heads, and \( d \) is the head dimension (feature size per head). In standard attention mechanisms, the computation cost grows quickly as the image gets bigger. Specifically, the complexity is:
\[
\mathcal{O}(2n^2hd)
\]
This means the model needs to do a lot of calculations as \( n \) increases.
\begin{figure}[h]
    \centering
    \includegraphics[width=8cm,height=2cm]{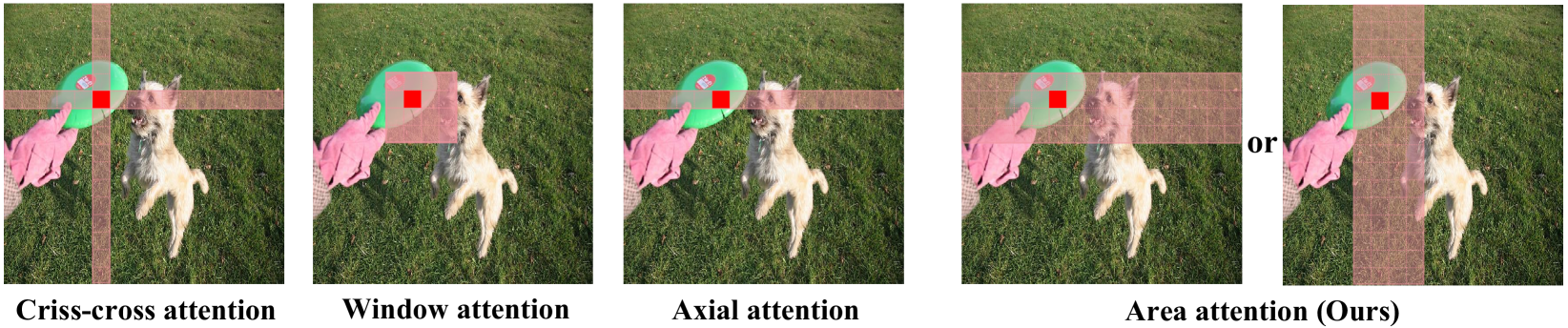}
    \caption{Area Attention uses a simple method to split the feature map into \( l \) equal parts, either vertically or horizontally (with \( l = 4 \) by default). This approach avoids complicated processing and still keeps a wide view of the image, making it fast and efficient~\cite{tian2502yolov12}.
}
    \label{Yolo_segment}
\end{figure}
\nix{
To fix this, YOLOv12 uses a smart method. Instead of looking at the entire image at once, it breaks the image into long horizontal or vertical stripes, called \textit{area segments}, as illustrated in figure \ref{Yolo_segment}. Each image has a height \( H \) and width \( W \). These segments have a size of either \( \left( \frac{H}{l}, W \right) \) or \( \left( H, \frac{W}{l} \right) \), and the default number of segments is \( l = 4 \).}

To solve this, YOLOv12 uses a clever approach. Instead of processing the entire image at once, it divides the image into long horizontal or vertical sections called \textit{area segments}, as shown in Figure~\ref{Yolo_segment}. Each image has a height \( H \) and a width \( W \). The feature map is split into segments of size either \( \left( \frac{H}{l}, W \right) \) or \( \left( H, \frac{W}{l} \right) \), where the default number of segments is \( l = 4 \). This method helps the model focus on smaller parts of the image while still working efficiently~\cite{tian2502yolov12}.
After applying attention within these smaller areas, the new complexity becomes:
\[
\mathcal{O}\left( \frac{1}{2} n^2 h d \right)
\]
This means it’s about twice as fast as the usual attention, but still powerful enough to understand spatial details in the image~\cite{tian2502yolov12}.
\section{Performance Metrics}\label{sec:metrics}

The evaluation metrics considered in this study include accuracy, sensitivity, specificity, precision, and F1-score, as defined in Equations~(1) to (5). These metrics are derived from the confusion matrix, which illustrates the relationship between actual and predicted classes. In this matrix, true positives (TP) and true negatives (TN) indicate correctly classified positive and negative instances, respectively, while false positives (FP) and false negatives (FN) represent misclassified negative and positive instances.
\begin{eqnarray}
  Accuracy &= & \frac{TP + TN}{TP + FP + TN + FN} ~~~~~~~\\
Sensitivity &=& \frac{TP}{TP + FN} \\
  Specificity &=& \frac{TN}{TN + FP} \\
  Precision &=& \frac{TP}{TP + FP} \\
  F1\text{-}score &=& \frac{2 \times Precision \times Recall}{Precision + Recall}~~~~~
\end{eqnarray}

\section{YOLOv12n Performance Results}\label{sec:YoloResults}

\subsection{Results of the Segmented Cell Images Using Hue Channel (YOLO)}

\begin{figure}[H]
    \centering
    \includegraphics[width=9cm,height=5cm]{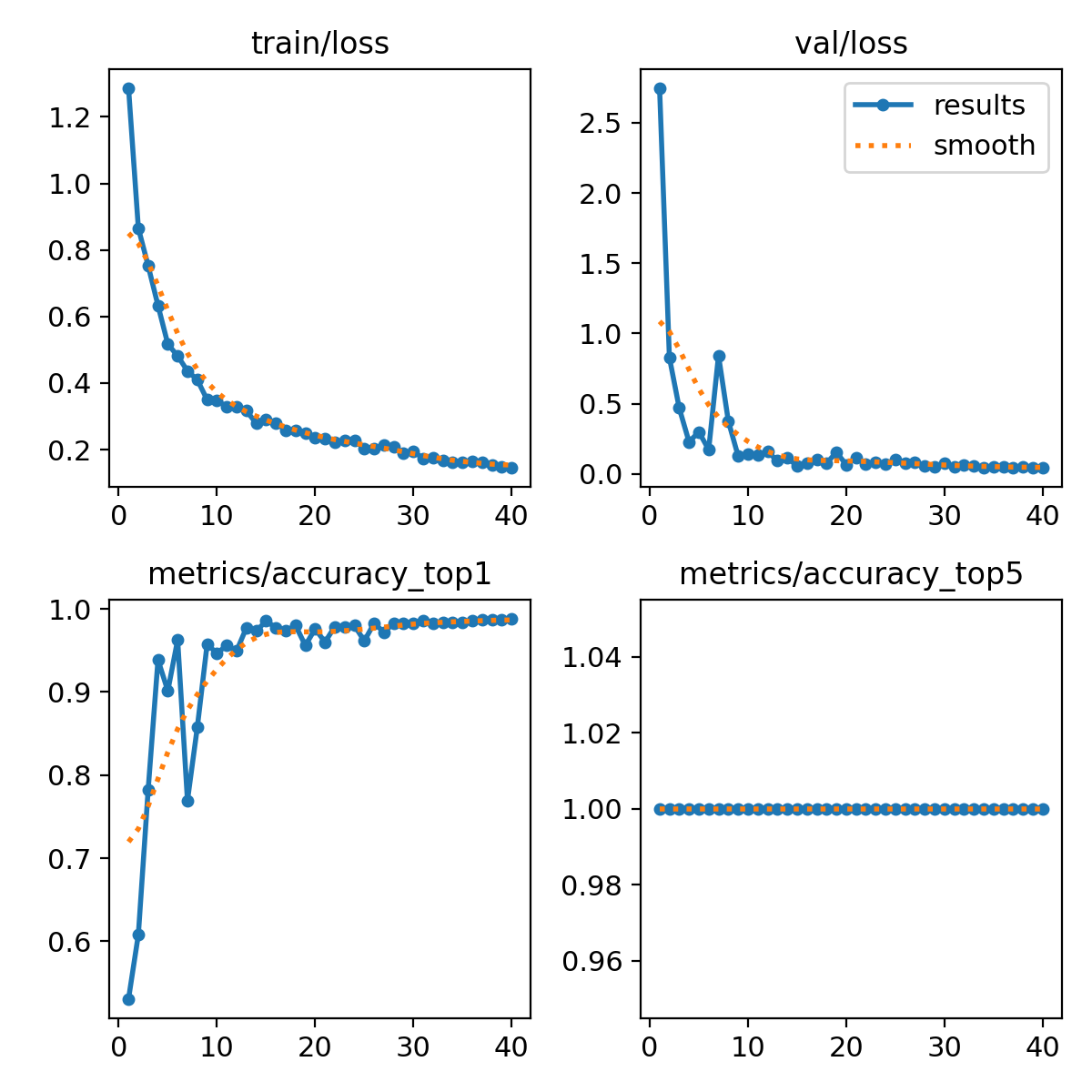}
    \caption{Training and validation loss curves for YOLO on hue-segmented cell images.}
    \label{hue_cell_loss_yolo}
\end{figure}

\begin{figure}[H]
    \centering
    \includegraphics[width=9cm,height=4cm]{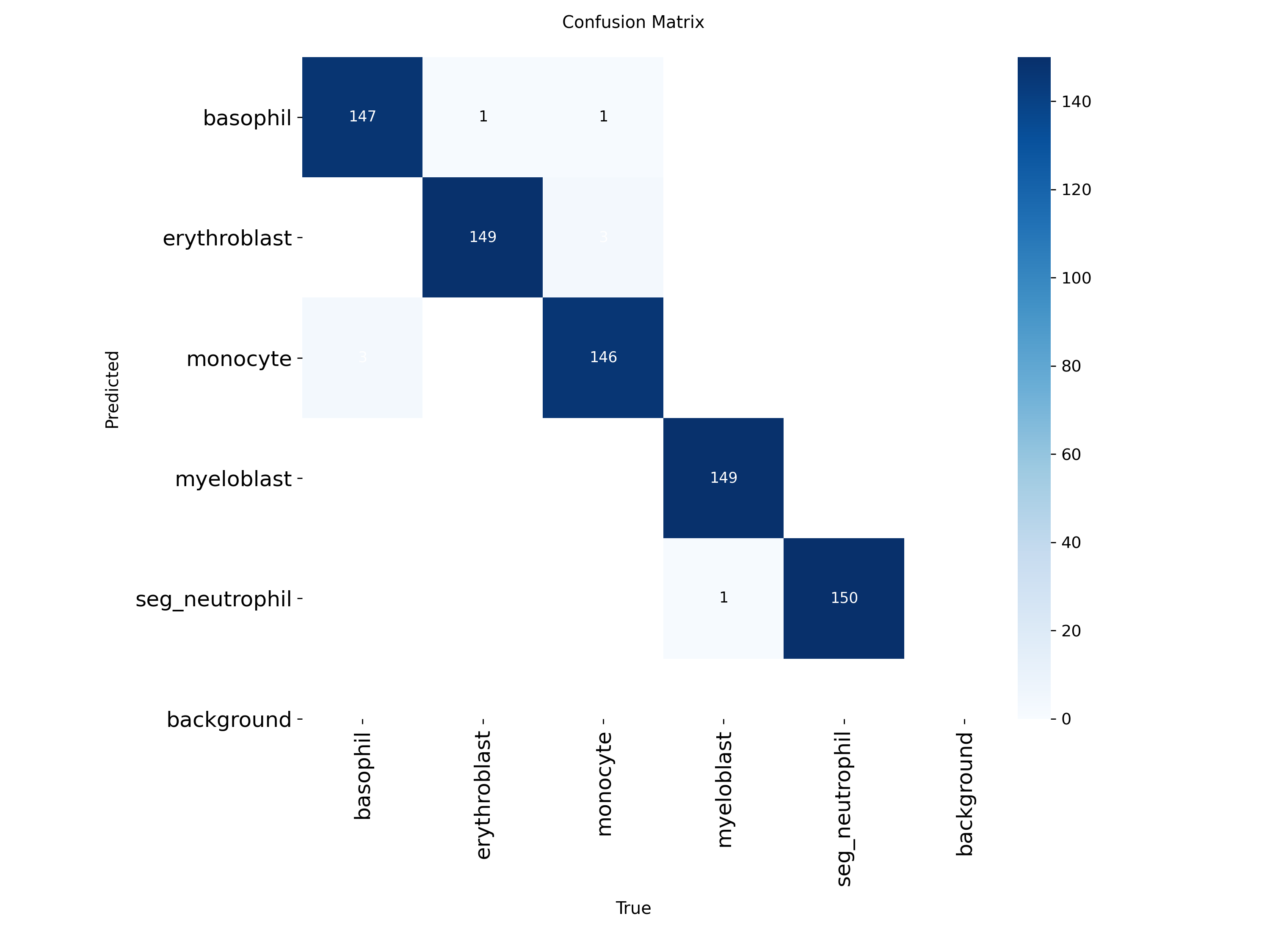}
    \caption{Confusion matrix of YOLO for hue-segmented cell image classification.}
    \label{hue_cell_confusion_yolo}
\end{figure}

In this section, we evaluate the performance of YOLOv12 models on hue-segmented cell images. The YOLOv12n model achieved a Validation and testing accuracy of 0.988, as illustrated in Figure~\ref{hue_cell_loss_yolo}. The training curve shows an overall improvement, with initial fluctuations that gradually stabilized as the number of epochs increased.
The training loss decreased steadily, reflecting effective learning, while the validation loss showed some early variability before trending downward. Furthermore, the confusion matrix in Figure~\ref{hue_cell_confusion_yolo} provides detailed insights into the model's classification performance across all cell types, demonstrating high precision, strong F1-scores, and consistent reliability in differentiating between classes.
\subsection{Results of the Segmented Cell Images Using Otsu Thresholding (YOLO)}
\begin{figure}[H]
    \centering
    \includegraphics[width=9cm,height=5cm]{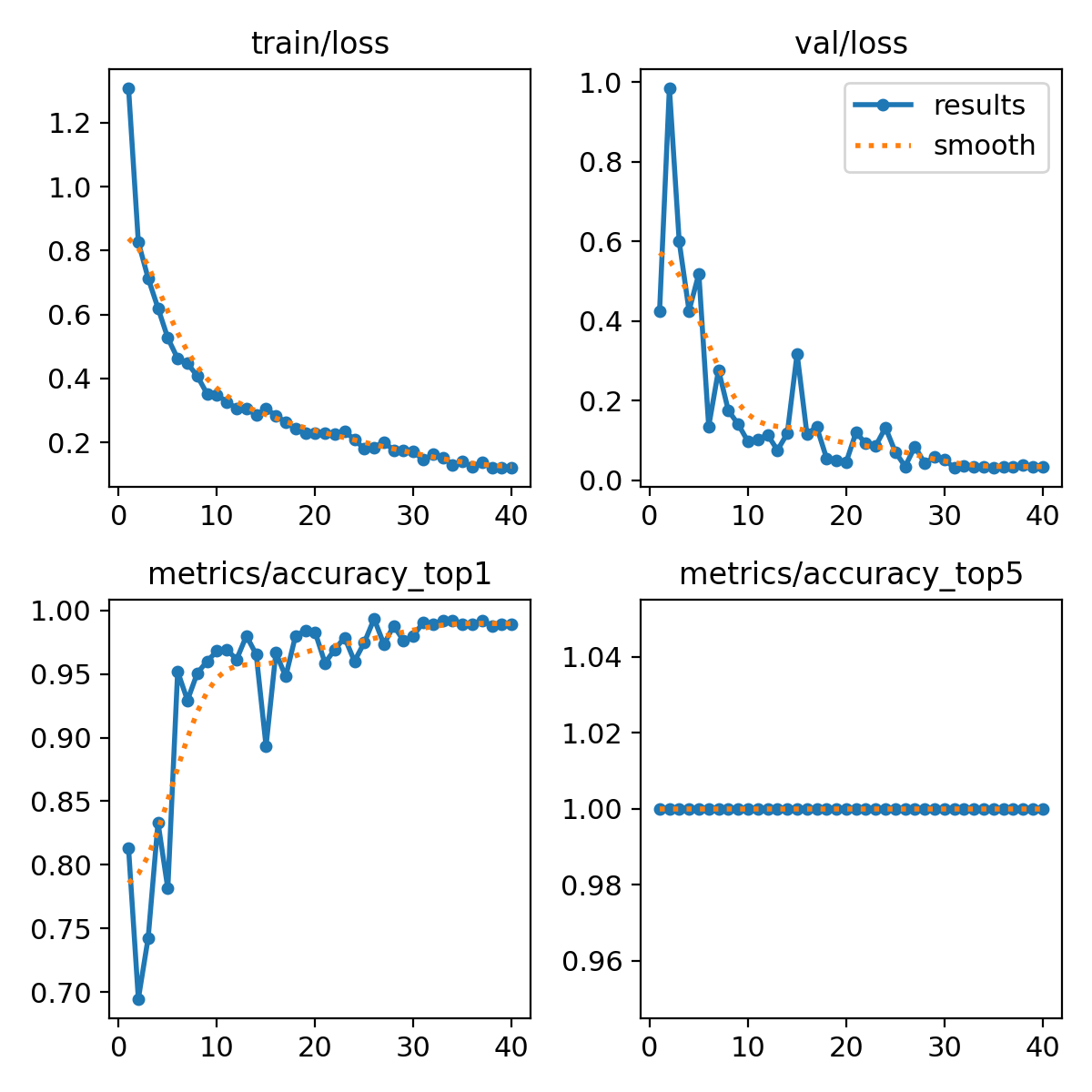}
    \caption{Training and validation loss curves for YOLO on Otsu-thresholded cell images.}
    \label{Cell_Otsu_loss_yolo}
\end{figure}

\begin{figure}[H]
    \centering
    \includegraphics[width=9cm,height=4cm]{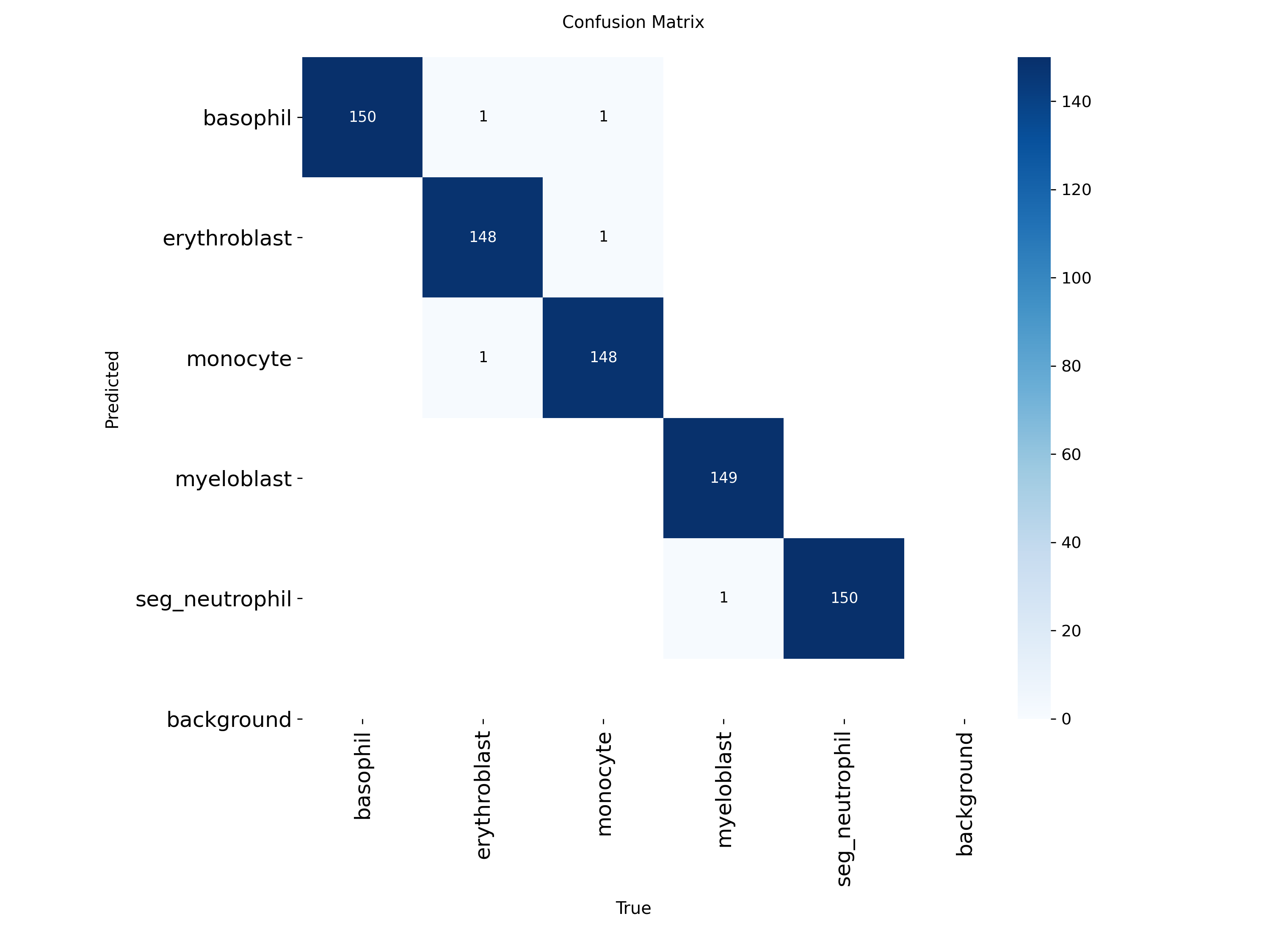}
    \caption{Confusion matrix of YOLO for Otsu-thresholded cell image classification.}
    \label{Cell_Otsu_confusion_yolo}
\end{figure}

This section presents the evaluation results of the YOLOv12 model trained on cell images segmented using Otsu thresholding. This method achieved the highest validation accuracy to date, reaching 0.993, and recorded an impressive test accuracy of 0.993 as well. As shown in Figure~\ref{Cell_Otsu_loss_yolo}, the loss curves reveal a degree of fluctuation during training, indicating some variability in learning, although the overall trend still reflects effective convergence.
Despite minor instabilities in the training loss, the model demonstrated strong generalization performance on the validation set. The confusion matrix in Figure~\ref{Cell_Otsu_confusion_yolo} further supports this, showing consistently high precision and reliable classification across all categories, confirming the robustness of this segmentation approach when paired with the YOLO architecture.
\subsection{Results of the Segmented Nucleus Images Using Hue Channel (YOLO)}
\begin{figure}[t]
    \centering
    \includegraphics[width=9cm,height=5cm]{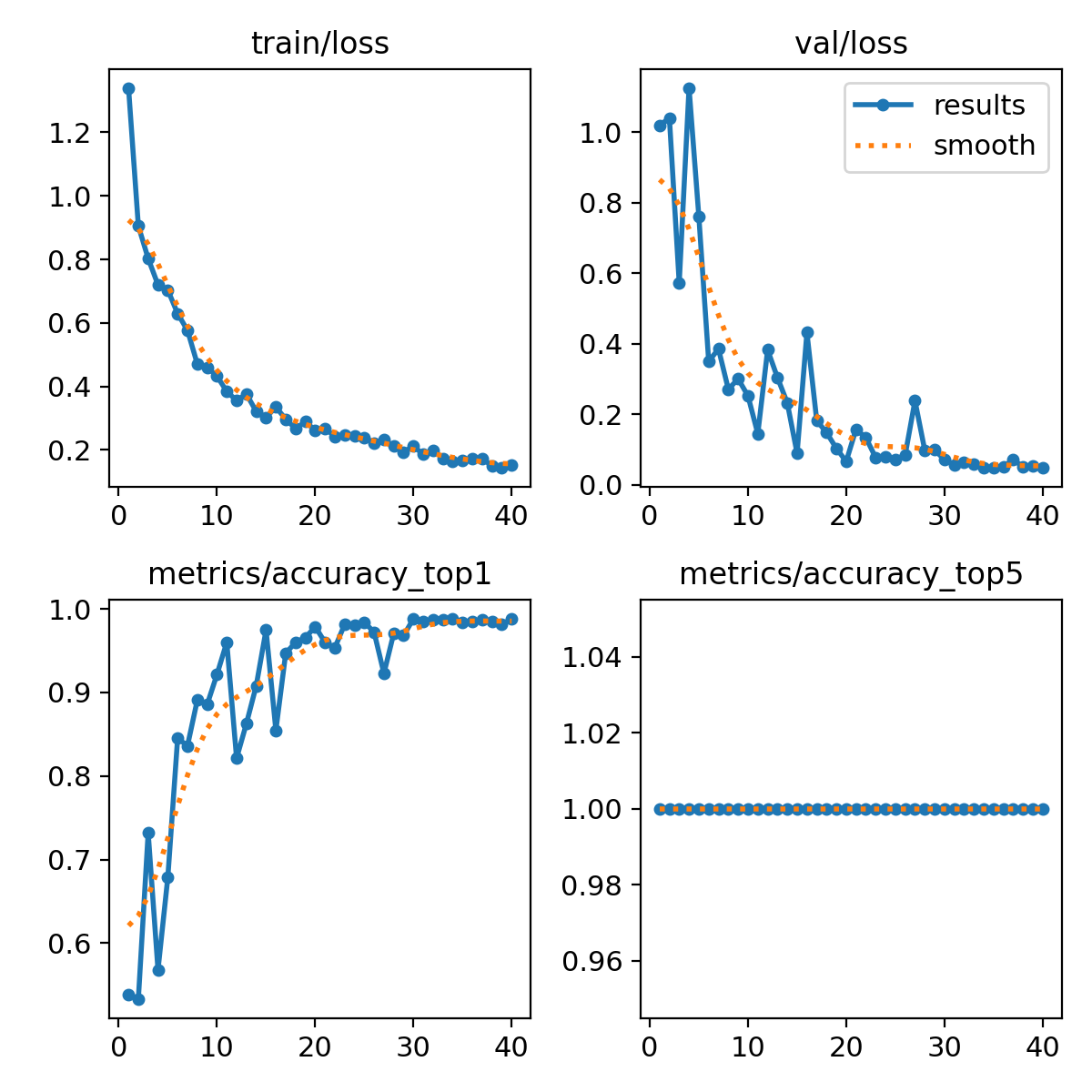}
    \caption{Training and validation loss curves for YOLO on hue-segmented nucleus images.}
    \label{Nucleus_Hue_loss_yolo}
\end{figure}

\begin{figure}[H]
    \centering
    \includegraphics[width=9cm,height=4cm]{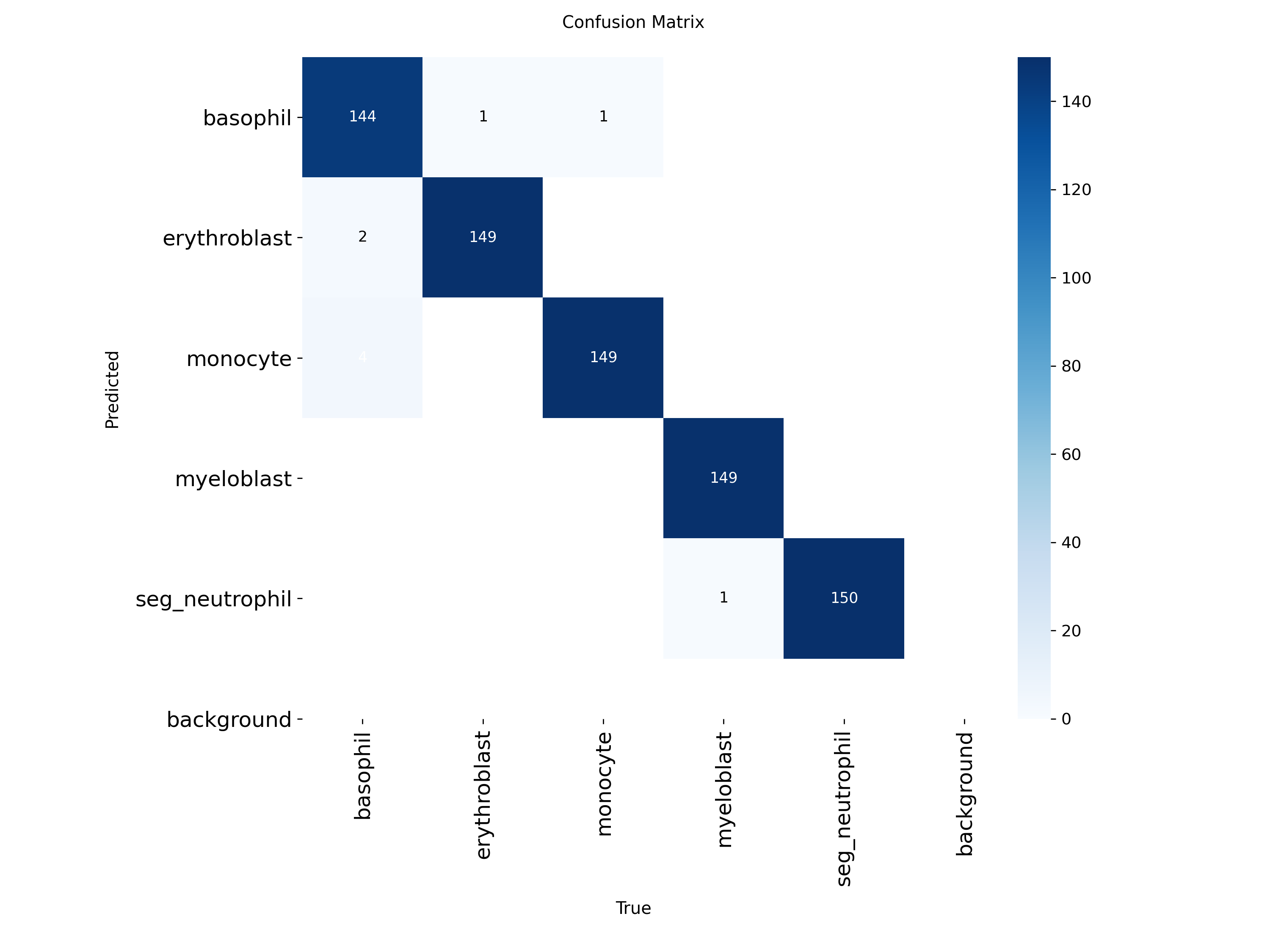}
    \caption{Confusion matrix of YOLO for hue-segmented nucleus image classification.}
    \label{Nucleus_Hue_Confusion_yolo}
\end{figure}

In this section, we evaluate the YOLOv12 model trained on nucleus images segmented using the hue channel. The model achieved a validation accuracy of 0.988 and a test accuracy of 0.988, indicating strong generalization capabilities. As illustrated in Figure~\ref{Nucleus_Hue_loss_yolo}, the training loss curve shows smooth and consistent convergence, demonstrating stable learning throughout the training process.
While the training curve was stable, the validation loss exhibited some fluctuations, suggesting minor inconsistencies in performance across epochs. However, these did not significantly affect the model's final classification ability. The confusion matrix shown in Figure~\ref{Nucleus_Hue_Confusion_yolo} further confirms the model’s high precision and accuracy across all five cell classes.

\subsection{Results of the Segmented Nucleus Images Using Otsu Thresholding (YOLO)}

This section presents the evaluation of the YOLOv12 model trained on nucleus images segmented via Otsu thresholding. The model attained a validation accuracy of 0.988 and a test accuracy of 0.988, demonstrating similar classification performance to the hue-based segmentation approach.

\begin{figure}[t]
    \centering
    \includegraphics[width=9cm,height=5cm]{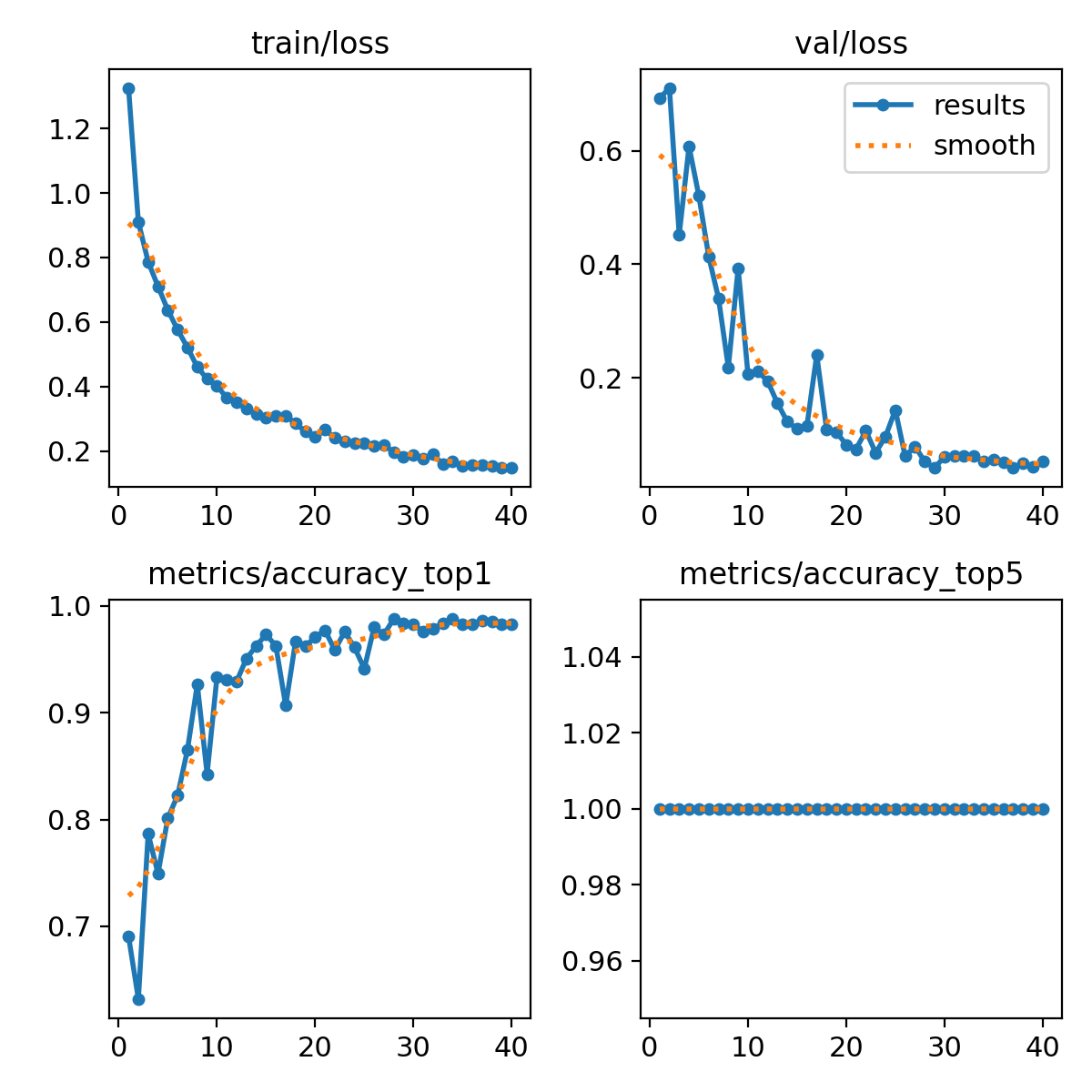}
    \caption{Training and validation loss curves for YOLO on Otsu-thresholded nucleus images.}
    \label{Nucleus_Otsu_loss_yolo}
\end{figure}

As shown in Figure~\ref{Nucleus_Otsu_loss_yolo}, the training loss exhibited a smooth and consistent decline, reflecting effective learning across epochs. The validation loss, however, displayed some fluctuations, similar to the hue-based segmentation results. These minor instabilities did not significantly impact the model’s overall performance.

\begin{figure}[h]
    \centering
    \includegraphics[width=9cm,height=4cm]{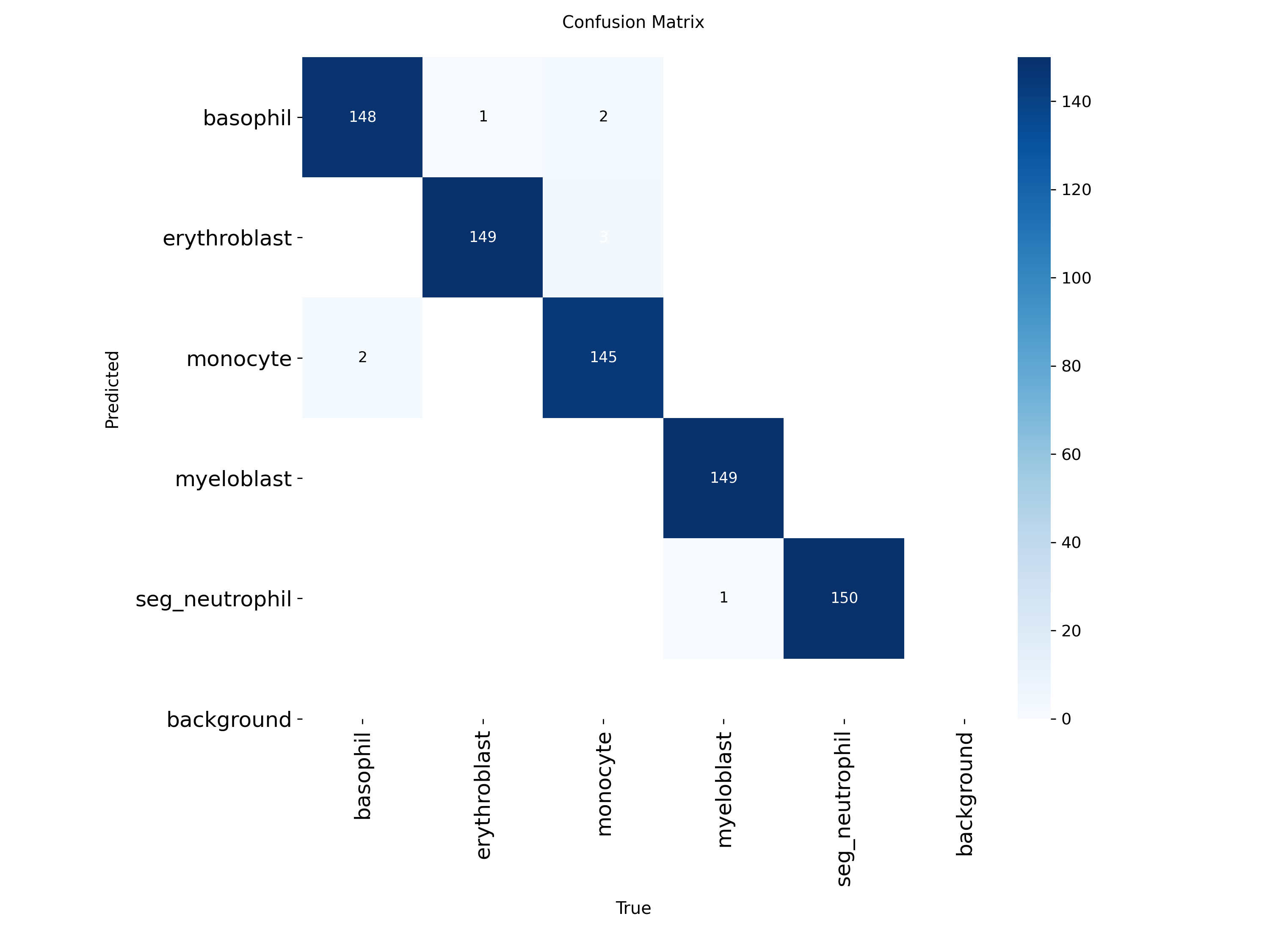}
    \caption{Confusion matrix of YOLO for Otsu-thresholded nucleus image classification.}
    \label{Nucleus_Otsu_confusion_yolo}
\end{figure}

The confusion matrix in Figure~\ref{Nucleus_Otsu_confusion_yolo} confirms the model's robustness and reliability across the five cell types, indicating strong agreement between predicted and actual labels.

Table~\ref{tab:yolo_accuracy_summary} summarizes the validation and test accuracies of YOLO models across different segmentation methods. Notably, the Otsu-thresholded cell images achieved the highest performance, with both validation and test accuracies reaching 0.993.

\begin{table}[h]
\centering
\caption{Summary of YOLO Accuracy Results}
\label{tab:yolo_accuracy_summary}
\begin{tabular}{@{}p{3.5cm}cc@{}}
\toprule
\textbf{Segmentation Method} & \textbf{Val Accuracy} & \textbf{Test Accuracy} \\
\midrule
Cell-Hue (YOLO) & 0.988 & 0.988 \\
Cell-Otsu (YOLO) & \textbf{0.993} & \textbf{0.993} \\
Nucleus-Hue (YOLO) & 0.988 & 0.988 \\
Nucleus-Otsu (YOLO) & 0.988 & 0.988 \\
\bottomrule
\end{tabular}
\end{table}
%
%

%
\subsection{Discussion}\label{}
The training curves across all models generally demonstrated stable convergence with smooth accuracy and loss trends. However, the YOLO models exhibited some fluctuations in the validation curves. Despite this, YOLO achieved excellent convergence and consistently outperformed the other models in terms of classification accuracy. In terms of performance, YOLOv12 emerged as the best-performing model, achieving the highest test accuracy of 99.3\% with Otsu-segmented cell images.  

\section{Conclusion}\label{sec:conclusion}
We addressed the problem of multiclass AML cell classification by implementing a complete pipeline involving image segmentation and deep learning-based classification.  We explored two segmentation strategies to create cell-based and nucleus-based segmented images. The two main segmentation methods were Hue Channel and Otsu thresholding. These methods were evaluated using ResNet50, Inception-ResNet50 v2, and YOLO v12 classification models. The models were trained and validated across five AML cell types: basophil, erythroblast, monocyte, myeloblast, and segmented neutrophil. Our findings show that YOLO v12 improved classification performance.

%
\begin{small}
\bibliographystyle{ieeetr}
\bibliography{references}
\end{small}

\end{document}